\documentclass[sigconf, review=false]{acmart}
\makeatletter
\def\@ACM@checkaffil{
    \if@ACM@instpresent\else
    \ClassWarningNoLine{\@classname}{No institution present for an affiliation}%
    \fi
    \if@ACM@citypresent\else
    \ClassWarningNoLine{\@classname}{No city present for an affiliation}%
    \fi
    \if@ACM@countrypresent\else
        \ClassWarningNoLine{\@classname}{No country present for an affiliation}%
    \fi
}
\makeatother

\usepackage{booktabs} 
\usepackage{tikz}
\usepackage{amsmath}
\usepackage{algorithm} 
\usepackage{algpseudocode}
\usepackage{balance}
\usepackage{flushend}
\setcopyright{rightsretained}

\acmConference[ICVGIP'24]{12th Indian Conference on Computer Vision, Graphics and Image Processing}{December 2024}{IIIT Bangalore, India}
\acmYear{2024}
\copyrightyear{2024}

\acmPrice{15.00}

\begin{document}
\title{Streamlining Video Analysis for Efficient Violence Detection}

\author{Gourang Pathak}
  \affiliation{%
    \institution{Vehant Technologies Pvt. Ltd.}
    \streetaddress{B-24, B Block, Sector 59}
    \postcode{201301}
  }
  
\author{Abhay Kumar}
  \affiliation{%
    \institution{Vehant Technologies Pvt. Ltd.}
    \streetaddress{B-24, B Block, Sector 59}
    \postcode{201301}
  }
  \author{Sannidhya Rawat}
  \affiliation{%
    \institution{Vehant Technologies Pvt. Ltd.}
    \streetaddress{B-24, B Block, Sector 59}
    \postcode{201301}
  }
  \author{Shikha Gupta}
  \affiliation{%
    \institution{Vehant Technologies Pvt. Ltd.}
    \streetaddress{B-24, B Block, Sector 59}
    \postcode{201301}
  }

\renewcommand{\shortauthors}{}

\begin{abstract}
This paper addresses the challenge of automated violence detection in video frames captured by surveillance cameras, specifically focusing on classifying scenes as "fight" or "non-fight." This task is critical for enhancing unmanned security systems, online content filtering, and related applications. We propose an approach using a 3D Convolutional Neural Network (3D CNN)-based model named X3D to tackle this problem. Our approach incorporates pre-processing steps such as tube extraction, volume cropping, and frame aggregation, combined with clustering techniques, to accurately localize and classify fight scenes. Extensive experimentation demonstrates the effectiveness of our method in distinguishing violent from non-violent events, providing valuable insights for advancing practical violence detection systems.
\end{abstract}

%
%



\maketitle

\section{Introduction}



In this work, we employ the X3D architecture \cite{Feichtenhofer2020X3D}, an efficient 3D convolutional neural network (CNN) specifically designed for video processing. X3D extends 2D image classification networks by progressively scaling their spatial and temporal dimensions, as well as increasing network width and depth. These enhancements make X3D particularly effective for handling the complexities of video data.  

Surveillance footage (CCTV) presents challenges such as high resolution, variable lighting, distorted perspectives, occlusions, and crowded scenes. These issues, combined with the challenge of localizing activity across the entire scene, complicate activity recognition. Real-time processing adds further complexity, demanding significant computational power.

\begin{figure}[h!] 
  \centering 
  \includegraphics[width=0.26\textwidth]{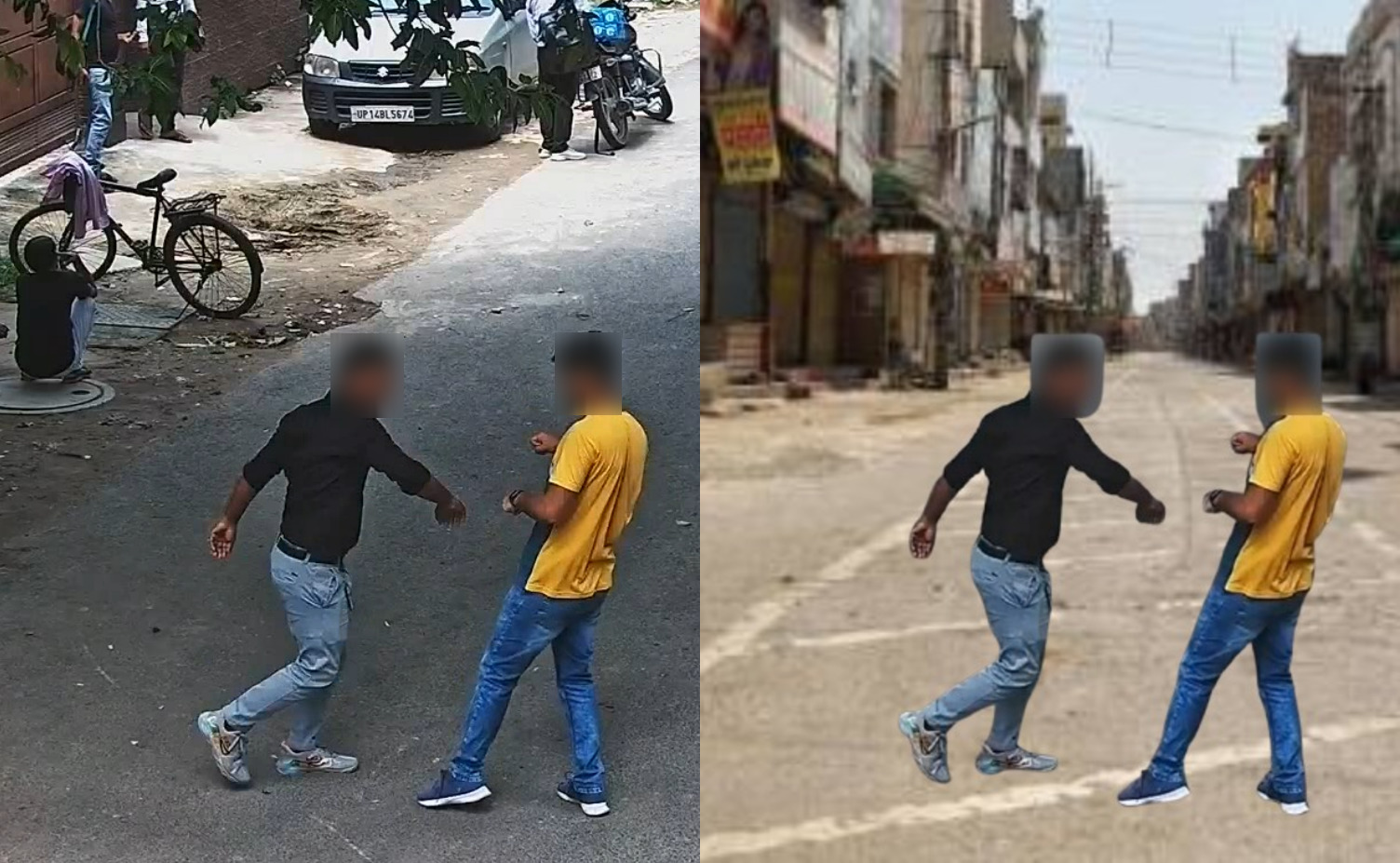} 
  \caption{Data Augmentation} 
  \vspace{-0.5em}
  \label{Figure 1} 
\end{figure}

Our key contributions are as follows:

\begin{enumerate} 

\item \textbf{Data Augmentation:} To enhance the model's accuracy, we implemented various data augmentation techniques Figure \ref{Figure 1}, including image segmentation with diverse background variations \cite{bgsub}. This increases the diversity of positive (fight) cases and improves model generalisation. 

\item \textbf{Localized Tube Extraction:} We localise fight scenes in videos by applying bounding box clustering and extracting cropped volumes. This approach offers greater robustness compared to the method in \cite{pytorchvideo_2024}. 

\item \textbf{Multiple Use Cases:} We design the model with flexibility, allowing it to easily adapt to various activity recognition tasks, such as detecting object-related violence, individuals collapsing, or object snatching. \end{enumerate}

\section{Dataset Preperation and Preprocess}

The X3D model \cite{fan2021pytorchvideo} uses a custom dataset loader, supporting input formats like npy, memmap, and h5 for training and inference. Based upon our observations, the time taken by npy format \cite{npy} had around a 100 times faster load time than memmap and around 20 times faster load time than h5 format, estimated on a machine with an NVIDIA GeForce RTX 3070 and 8192 MB of memory.


\begin{figure}[h!] 
  \centering 
  \includegraphics[width=0.4\textwidth]{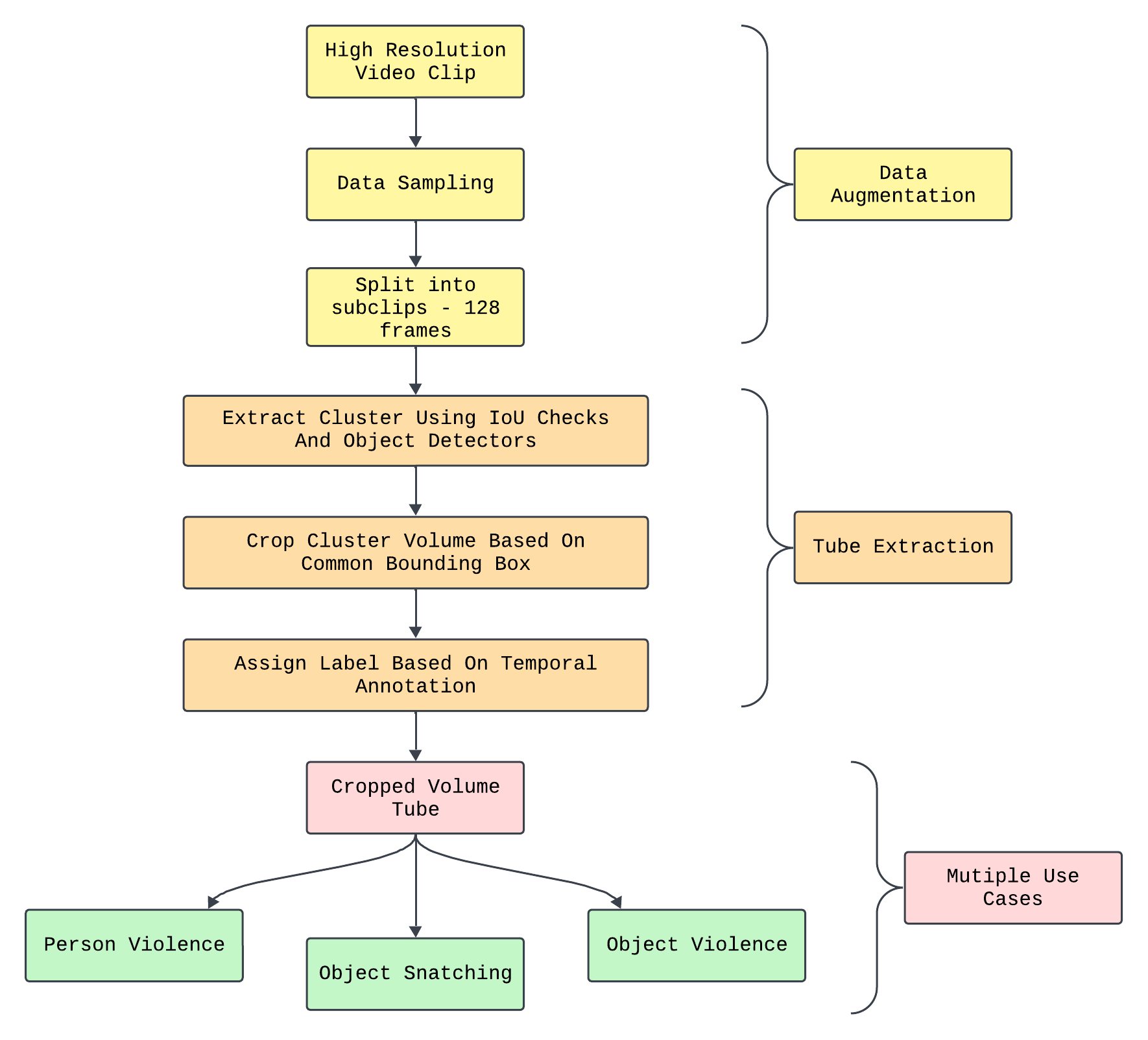} 
  \caption{Framework for tube extraction and its related applications} 
  \vspace{-0.5em}
  \label{Figure 2} 
\end{figure}
The overall framework for tube extraction and related applications appears in Figure \ref{Figure 2}. We process the input video by generating non-overlapping 128-frame clips, labelling each as either a fight or non-fight scenario, and resizing them to a standardised input size before training the model. We train the model using multiple violence detection datasets, including RWF-2000, the Fight Detection Surveillance dataset, and staged enactments. In our recorded and data-augmented enactments, we included 630 training cases (324 fight, 306 non-fight), 91 test cases (47 fight, 44 non-fight), and 179 validation cases (91 fight, 88 non-fight).

\begin{figure}[h!] 
  \centering 
  \includegraphics[width=0.4\textwidth]{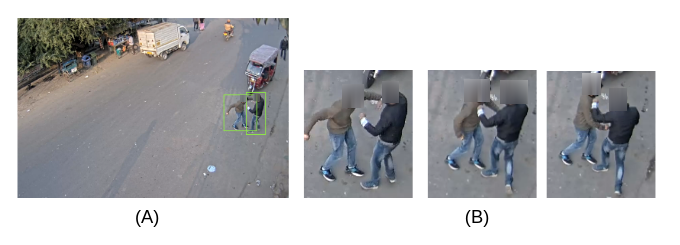} 
  \caption{Extracted Fight Tube} 
  \label{Figure 3} 
\end{figure}

To localise fight scenes, we cluster the detected persons' bounding boxes using Intersection over Union (IoU) checks and identify a common bounding box across frames. We simplify the method from \cite{Dave2022GabriellaV2:}, which relies on tracking IDs. Since tracking often fails in occlusion or low-light conditions, we address these challenges by focussing on true detections. Figure \ref{Figure 3}A shows the main video frame with overlapping boxes, and Figure \ref{Figure 3}B displays the extracted tubes. We detail the process steps in Algorithm \ref{alg:clustering_temporal}.

\begin{algorithm}
\caption{Steps for Video Tube Generation }
\label{alg:clustering_temporal}
\begin{algorithmic}[1]
\State \textbf{Input:} 
Video $V$ with $F$ frames, each frame $f \in V$ of size $(H, W, C)$, and a label vector \( T \) of size \( F \) for Fight/Non-Fight per frame.
\State \textbf{Output:} Resized video segments with person clusters $(128, 224, 224, 3)$ and Fight/Non-Fight Labels.

\State \textbf{Step 1:} Segment the video $V$ into smaller 128-frame volumes $\{V_i\}_{i=1}^{n}$ where each $V_i \subseteq V$ contains 128 consecutive frames.

\For{each volume $V_i$}
    \For{each frame $f_j \in V_i$}
        \State Detect persons in $f_j$ using standard object detector, yielding bounding boxes $B_j = \{b_1, b_2, \dots, b_k\}$ for $k$ detected persons.
    \EndFor
    
    \State \textbf{Step 2:} Retrieve the temporal annotation from $Z$ for $V_i$.
    \State Check if more than 70\% of the frames in $T$ for $V_i$ are labeled as fight.
    \If{fight frames $> 70\%$}
        \State Label the segment $V_i$ as a fight segment.
    \Else
        \State Label the segment $V_i$ as a non-fight segment.
    \EndIf

    \State \textbf{Step 3:} For each frame $f_j$, compute Intersection over Union (IoU) for all pairs of bounding boxes $b_i, b_{i'} \in B_j$.
    \State \textbf{Step 4:} If IoU$(b_i, b_{i'}) \geq \text{threshold}$, assign the corresponding persons to the same cluster $c_l$.
    
    \For{each cluster $c_l$ across all frames in $V_i$}
        \State \textbf{Step 5:} Compute the best bounding box $B_l$ for cluster $c_l$ by taking:
        \[
        B_l = \text{min}(x_1, y_1), \ \text{max}(x_2, y_2)
        \]
        across all frames in $V_i$ that belong to cluster $c_l$., where \text{$(x_1, y_1)$} and \text{$(x_2, y_2)$} is top-left and bottom-right points corresponding to cluster $c_l$ in each frame.
        
        \State \textbf{Step 6:} Extract cropped frames using the calculated bounding box $B_l$ for all frames in the volume $V_i$.
    \EndFor
    
    \State \textbf{Step 7:} Resize each extracted tube (set of 128 frames per cluster) to $(128, 224, 224, 3)$ assign the tube the corresponding Fight/Non-Fight Label.
\EndFor

\end{algorithmic}
\end{algorithm}


\section{Results and Analysis}


The model used StepLR and Cosine Annealing \cite{cosine} schedulers to optimize learning rates, enhancing accuracy and limiting overfitting, with a 70-10-20 split for training, testing, and validation.





    
    
    

\noindent
The following metrics were calculated for the model performance:
\begin{itemize}
    \item \textbf{Accuracy:} 0.86
    \item \textbf{Precision:} 0.87
    \item \textbf{Sensitivity:} 0.87
    \item \textbf{Specificity:} 0.82
\end{itemize}
\noindent

\begin{table}[h!]
    \centering
    \begin{tabular}{|c|c|}
    \hline
    \textbf{Dataset} & \textbf{Accuracy in \%} \\
    \hline
    RWF-2000 & 84.02 \\
    \hline
    VioPeru & 62.96 \\
    \hline
    Our Enactments & 95.6 \\
    \hline
    \end{tabular}
    \vspace{0.22cm}
    \caption{Accuracy on standard and our datasets}
    \label{table-2}
\end{table}

\vspace{-6mm}

Table \ref{table-2} summarizes the model's performance across datasets. The RWF-2000 test split included 116 fight and 78 non-fight cases, while the Vioperu dataset had  17 fight and 10 non-fight cases \cite{ncbi_2024}. Our enactments included 47 fight and 44 non-fight cases. Previous work \cite{alpa} achieved 84.5\% accuracy on RGB videos using background suppression, though this approach is error-prone. For the Vioperu dataset, model accuracy declined due to low resolution, inconsistent annotations, and limited diversity.

\section{Conclusions}
The challenge of detecting violent behaviours in surveillance footage has been addressed through our innovative approach. We utilised data augmentation techniques to enhance positive instances of fight data and employed cropped volume tubes for precise localisation of fight and non-fight scenes. These cropped tubes lay the groundwork for developing additional models to identify behaviours such as object violence, person collapse, and object snatching. Achieving an overall accuracy of 86\% in violence detection, our model demonstrates strong performance. Its adaptability to diverse scenarios highlights its potential for broader applications in behaviour analysis and surveillance systems.



\bibliographystyle{ACM-Reference-Format}
\newpage
\bibliography{action}

\end{document}